\RequirePackage{booktabs}

\documentclass[sn-mathphys-num,iicol]{sn-jnl}

\usepackage[caption=false, font=footnotesize]{subfig}
\usepackage{graphicx}%
\usepackage{multirow}%
\usepackage{amsmath,amssymb,amsfonts}%
\usepackage{amsthm}%
\usepackage{mathrsfs}%
\usepackage[title]{appendix}%
\usepackage[table]{xcolor}%
\usepackage{textcomp}%
\usepackage{manyfoot}%
\usepackage{booktabs}%
\usepackage{listings}%

\usepackage{csquotes}
\usepackage{makecell}
\usepackage[ruled, vlined]{algorithm2e}

\newcommand\blfootnote[1]{%
  \begingroup
  \renewcommand\thefootnote{}\footnote{#1}%
  \addtocounter{footnote}{-1}%
  \endgroup
}

\begin{document}

\title[Future-Proofing Class-Incremental Learning]{Future-Proofing Class-Incremental Learning}

\author*[1,2]{\fnm{Quentin} \sur{Jodelet}}\email{jodelet@net.c.titech.ac.jp}

\author[2]{\fnm{Xin} \sur{Liu}}\email{xin.liu@aist.go.jp}

\author[1]{\fnm{Yin Jun} \sur{Phua}}\email{phua@c.titech.ac.jp}

\author[1,2]{\fnm{Tsuyoshi} \sur{Murata}}\email{murata@c.titech.ac.jp}

\affil*[1]{\orgdiv{Department of Computer Science}, \orgname{Tokyo Institute of Technology}, \orgaddress{\street{W8-59 2-12-1 Ookayama}, \city{Meguro}, \postcode{152-8552}, \state{Tokyo}, \country{Japan}}}

\affil[2]{\orgdiv{Artificial Intelligence Research Center}, \orgname{AIST}, \orgaddress{\street{2-4-7 Aomi}, \city{Koto}, \postcode{135-0064}, \state{Tokyo}, \country{Japan}}}

\abstract{Exemplar-Free Class Incremental Learning is a highly challenging setting where replay memory is unavailable. Methods relying on frozen feature extractors have drawn attention recently in this setting due to their impressive performances and lower computational costs. However, those methods are highly dependent on the data used to train the feature extractor and may struggle when an insufficient amount of classes are available during the first incremental step. To overcome this limitation, we propose to use a pre-trained text-to-image diffusion model in order to generate synthetic images of future classes and use them to train the feature extractor. Experiments on the standard benchmarks CIFAR100 and ImageNet-Subset demonstrate that our proposed method can be used to improve state-of-the-art methods for exemplar-free class incremental learning, especially in the most difficult settings where the first incremental step only contains few classes. Moreover, we show that using synthetic samples of future classes achieves higher performance than using real data from different classes, paving the way for better and less costly pre-training methods for incremental learning.}

\keywords{Class Incremental Learning, Continual Learning, Image Classification, Image Generation}

\maketitle

\section{Introduction}
\label{section:introduction}

\blfootnote{The version of record of this article, first published in \enquote{Machine Vision and Applications}, is available online at Publisher's website: \url{https://doi.org/10.1007/s00138-024-01635-y}}

While  humans naturally learn and accumulate knowledge continuously during their whole lifespan, deep neural networks struggle to do the same. Continuously training deep neural networks on new data will inevitably induce a severe decrease of the performance on the previously learned data. This phenomenon, named catastrophic forgetting~\citep{MCCLOSKEY1989109,ratcliff1990connectionist}, has received more and more attention during the past decade, and mitigating it is the core problem of incremental learning and continual learning. Among the different settings of incremental learning, Class-Incremental Learning (CIL) is among the most challenging ones. In this setting, the model is incrementally trained on a few number of classes at a time, without having access to the previously learned ones. The objective is to train a unified model that performs well on every encountered class.

Rehearsal memory~\citep{Rebuffi_2017_CVPR,chaudhry2019tiny,Lomonaco_2022} is one of the most commonly used methods to mitigate catastrophic forgetting and is used by most of the approaches for Class-Incremental Learning. It consists of a small buffer containing few representative exemplars of each previously learned class that are later used for replay while learning new classes. However, there remain situations where using a rehearsal memory is not possible, either due to privacy concerns regarding the training data (especially for tasks such as person re-identification) or due to limitations in the storage space. This led to the emergence of a new setting named Exemplar-Free Class-Incremental Learning (EFCIL) or Class-Incremental Learning Without Memory. Methods for EFCIL often freeze the feature extractor after the first incremental step~\citep{hayes2020lifelong,Petit_2023_WACV}, achieving competitive performances while allowing for a faster and less computationally expensive learning of new classes compared to methods that have to backpropagate through the entire deep neural network. However, this may also lead to poor performance when the number of classes available in the initial step to train the feature extractor is too limited.

To address this limitation, several works have proposed to pre-train the feature extractor \citep{Lomonaco_2022,zhou2024continual} using additional data from classes similar yet different from those in the target dataset. However, with the advent of foundational diffusion model for image generation, pre-training the feature extractor on a more targeted dataset, specifically classes that are highly likely to be encountered during the future incremental steps, has the potential to lead to better improvements in this setting. While the aspect of using future classes to pre-train the model beforehand seems counter to the concept of Class-Incremental Learning, we argue that it is indeed a realistic setting. In many real world applications, collecting training data for every possible object in an ever changing world to train an omniscient classifier is impossible. It is much more realistic to train a classifier that is more specific to the needs of the user. Even under such a scenario, collecting every possible object that the user needs is an insurmountable task. Therefore, it is much more desirable to attempt to predict specific objects that are likely to be encountered, pre-train the feature extractor, and then perform Incremental Learning while it is deployed to the user. In our case, using a diffusion model that generates training data makes it possible to pre-train especially when objects can be difficult or impossible to obtain.

In this work, we propose to leverage pre-trained text-to-image diffusion models to better prepare Exemplar-Free Class-Incremental Learning methods relying on a frozen feature extractor for the future. The diffusion models generate synthetic samples of possible future classes that will be jointly used with actual real data during the first incremental step to train a feature extractor that better generalizes to future classes. Experiments using various challenging settings and datasets for EFCIL highlight how our method can be combined with state-of-the-art approaches and results in significant improvement of the average incremental accuracy. Moreover, it appears that using synthetic images of future classes achieves higher performance than using real images from different yet similar classes. 

Our contributions can be summarized into three points:
\begin{itemize}
    \item We propose a framework relying on pre-trained diffusion models to better prepare Exemplar-Free Class-Incremental Learning methods for the future, reaching new state-of-the-art performances.
    \item While previous approaches for Incremental Learning used pre-trained diffusion models for generating samples from past classes, our proposed method aims at generating samples from future classes instead.
    \item We experimentally show that using synthetic samples of future classes for training the feature extractor achieves higher performance than using real images of classes different from those in the target dataset.
\end{itemize}

\section{Related Work}

\subsection{Class-Incremental Learning (CIL) and Exemplar-Free Class-Incremental Learning (EFCIL)}
While Class-Incremental Learning may be applied to various tasks and settings such as semantic segmentation~\citep{Douillard_2021_CVPR,Cong_2024} or federated learning~\citep{Dong_2022,Dong_cvpr_2023}, in this work we only consider the image classification task which is the most extensively studied.
Most approaches for Class-Incremental Learning~\citep{BELOUADAH202138,zhou2023deep} leverage knowledge distillation~\citep{Rebuffi_2017_CVPR,Hou_2019_CVPR,douillard2020podnet,Dong_2023}, a rehearsal memory and a method to mitigate the bias toward the new classes~\citep{Hou_2019_CVPR,wu2019large,10.1007/978-3-030-86340-1_31,Ahn_2021_ICCV} induced by the limited size of the replay memory. Recent works~\citep{yan2021dynamically,zhou2023model} proposed to extend the feature extractor at each incremental step to learn the new classes while preserving past knowledge by freezing previously learned weights.

Exemplar-Free Class-Incremental Learning is a more challenging setting in which there is no rehearsal memory, making the replay of samples from the previously encountered classes impossible. Similarly to approaches designed for Class-Incremental Learning, numerous methods for Exemplar-Free Class-Incremental Learning also continuously finetune the feature extractor and the classifier~\citep{Wu_2021_ICCV,zhu2021class,JODELET2022103582,Zhu_2021_CVPR,zhu2022self}. Several authors have highlighted the risk of overfitting on the current task and the importance of having general and transferable feature representation for EFCIL. To address this concern, approaches often rely on self-supervised learning~\citep{Wu_2021_ICCV,Zhu_2021_CVPR} or by creating augmented classes using Mixup~\citep{zhang2017mixup}. Recently, methods relying on a fixed feature extractor have gained more attention~\citep{hayes2020lifelong,ostapenko2022continual}. Those methods either rely on a pre-trained model~\citep{wang2022dualprompt,Wang_2022,Smith_2023,zhou2024continual} or by freezing the feature extractor after learning it during the initial step~\citep{hayes2020lifelong,Petit_2023_WACV,goswami2023fecam}. \citet{Petit_2023_WACV} proposed FeTrIL which achieves state-of-the-art performances by relying on a frozen feature extractor and a linear classifier trained using actual features of new classes and pseudo-features of past classes generated by a geometric translation of new class features. FeCAM~\citep{goswami2023fecam} is a recently proposed approach for Exemplar-Free Class-Incremental Learning with a fixed feature extractor that rely on the Mahalanobis distance combined with Tukey’s transformation, covariance shrinkage, and correlation normalization. Those methods exhibit numerous advantages including a less computationally expensive training procedure that allows for for a deployment on edge devices. However, they also depend on the initial training strategy used~\citep{janson2022a,Petit_2024_WACV} and may achieve poor performance if the quantity of data available for training the feature extractor during the initial step is too limited.

\subsection{Training on synthetic images}
While few attempts have been made using GAN-based models~\citep{besnier2020dataset,jahanian2022generative}, the use of synthetic images for training deep learning models has recently gained momentum following the advances in diffusion models~\citep{sohldickstein2015deep,ho2020denoising} and their wide availability. Diffusion models generate synthetic samples using an iterative denoising process conditioned on an input, usually a text prompt. Imagen~\citep{saharia2022photorealistic} performs the denoising task in the pixel space while Stable Diffusion~\citep{Rombach_2022_CVPR} and DALL-E2~\citep{ramesh2022hierarchical} perform it in the latent space in order to decrease the computational cost of the generative process. In one of the pioneer works, \citet{Sariyildiz_2023_CVPR} studied the use of the pre-trained text-to-image diffusion model Stable Diffusion to generate a synthetic replacement for the original training dataset. Their experiments highlighted that while the model trained using the real train dataset reaches a higher accuracy on the real test dataset compared to the one trained on synthetic images, they perform on par for transfer learning tasks. Concurrently, \citet{zhang2024expanding} proposed to expand real datasets using synthetic samples generated with various diffusion models and achieved significant improvement on various image classification datasets. Several authors proposed to use synthetic images generated by pre-trained diffusion models for different problems of computer vision such as zero-shot and few-shot learning~\citep{he2022synthetic}, long-tailed image classification~\citep{shin2023fill}, incremental learning~\citep{Jodelet_2023_ICCV,duan2023promptbased}, out-of-distribution detection~\citep{du2024dream}, or semantic segmentation~\citep{nguyen2024dataset}. \citet{tian2023stablerep} showed that the representation learned using contrastive learning on synthetic images generated by Stable Diffusion outperforms state-of-the-art methods trained on real datasets.

\subsection{Additional data for Incremental Learning}
Several authors~\citep{lee2019overcoming,zhang2020class,lechat2021pseudo,bellitto2022effects,liu2023online} have considered the use of additional data for Class-Incremental Learning by leveraging datasets of real images belonging to classes different from the ones that the model has to learn. Most of these approaches assume that the additional dataset is unlabeled, restricting the use of those additional data samples to the distillation loss~\citep{lee2019overcoming,zhang2020class,liu2023online}. \citet{Jodelet_2023_ICCV} first proposed to use large pre-trained text-to-image diffusion models for Incremental Learning in order to generate additional synthetic images for the classes previously encountered by the model. Using a textual prompt based on the class name, this method generates labeled samples that can be used for the distillation loss but also for any supervised losses such as the classification loss. More recently, pre-trained diffusion models were also used for Class-Incremental semantic segmentation~\citep{chen2023diffusepast} and object detection~\citep{kim2024sddgr}. However, there remains a significant distribution gap between the real samples and the synthetic ones~\citep{Jodelet_2023_ICCV} whose impact on the classifier makes the use of pre-trained diffusion model for replaying samples from past classes especially difficult for Exemplar-Free Class-Incremental Learning.

\subsection{Preparing for the future}
While most of the methods for Class-Incremental Learning only focus on how to prevent the forgetting of past data, some authors have studied how to better prepare the model for future updates. \citet{Douillard_2021} took inspiration from zero-shot learning and proposed to incorporate into the classifier feature vectors of unseen future classes during training. Those possible future vectors, named ghost features, are produced by a generator using actual features of seen classes and prior information about all the classes. \citep{pernici2021class} proposed to use a pre-allocated Regular Polytope Classifier which can exploit future unseen classes as negative examples. \citet{zhou2022forward} proposed to reserve embedding space for future classes by pre-assigning virtual prototypes and by forecasting future classes using manifold Mixup~\citep{zhang2017mixup}. \citet{bellitto2022effects} proposed to use an auxiliary dataset of real images containing classes unrelated to the main task in order to prepare the model for the future. Those additional data samples are used to learn more general and stable representation that can be leveraged when learning future classes.

\subsection*{}
Compared to previously mentioned methods, in this work, we aim to leverage pre-trained text-to-image diffusion models not for generating additional past data but for generating synthetic samples for possible future classes. We study how those synthetic images of future classes can be used during the initial incremental step and to what extent they can help to better prepare Exemplar-Free incremental learning methods relying on a frozen feature extractor.

\section{Proposed method}

The objective of Class-Incremental Learning (CIL) is to learn a unified model $\theta$ on a gradually increasing set of classes. $\theta$ can be decomposed into $\{\Phi,\mathcal{W} \}$ with $\Phi$ the feature extractor and $\mathcal{W}$ the classifier. The training procedure is divided into $T$ incremental steps, each step consisting of a dataset $\mathcal{D}_t$ containing samples from the set of new classes $\mathcal{Y}_t$ such that $\forall i,j \in \{1,2,...,T\}, \mathcal{Y}_i \cap \mathcal{Y}_j = \emptyset$ for $i \neq j$. Contrary to standard Class-Incremental Learning, in the Exemplar-Free Class-Incremental Learning (EFCIL) setting, the model does not have access to any replay memory: during each incremental step $t$, the model $\theta_t$ is trained solely on the current dataset $\mathcal{D}_t$.
After each incremental step, the model is evaluated on the test dataset of all the classes observed so far without having any step descriptor.

\subsection{Future-Proofing Class-Incremental Learning}

\begin{figure*}
\begin{center}
\resizebox{\textwidth}{!}{%
\includegraphics{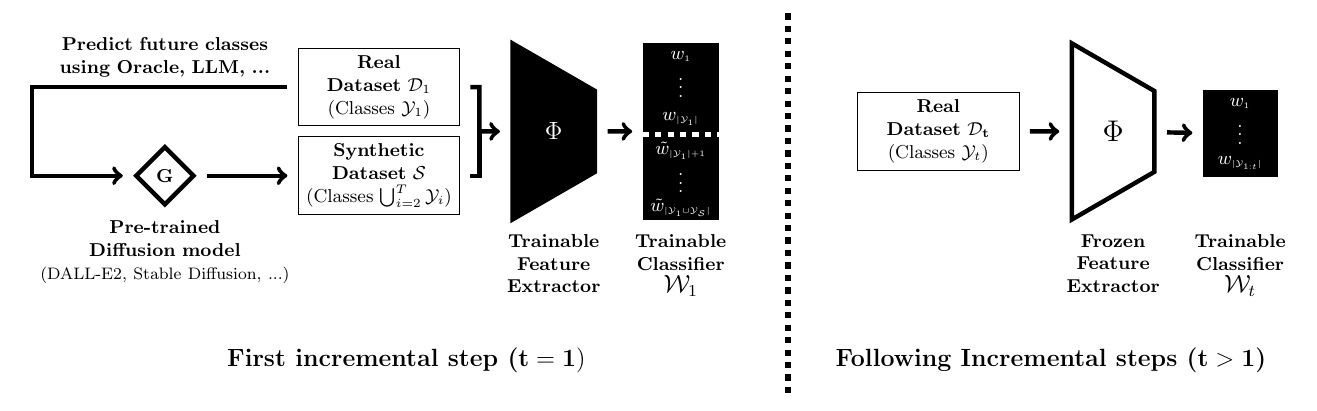}}
\end{center}
   \caption{Diagram representing our proposed method \textbf{F}uture-\textbf{P}roof \textbf{C}lass \textbf{I}ncremental \textbf{L}earning (\textbf{FPCIL}) which prepares the feature extractor for the future by jointly training it on an auxiliary dataset in addition to the current dataset during the initial step. The auxiliary dataset contains synthetic samples of future classes generated using a pre-trained text-to-image diffusion model. After the first incremental step, the feature extractor $\phi$ is frozen and the weights of the classifier $\tilde{w}$ corresponding to the future classes are removed. Afterward, in the following incremental step, only the classifier $\mathcal{W}$ is trained using dedicated method (e.g. FeTrIL, FeCAM).}
\label{fig:diagramfpcil}
\end{figure*}

In order to create a future-proof model, as illustrated in Figure~\ref{fig:diagramfpcil}, we introduce \textbf{F}uture-\textbf{P}roofing \textbf{C}lass \textbf{I}ncremental \textbf{L}earning (\textbf{FPCIL}), a straightforward approach which can be combined with any method for Class-Incremental Learning relying on a frozen feature extractor. During the first incremental step, when training the feature extractor, in addition to the initial step dataset $\mathcal{D}_1$, an additional dataset $\mathcal{S}$ containing $\lvert \mathcal{Y}_\mathcal{S} \rvert$ different classes is used. 
While any labeled datasets could be used for $\mathcal{S}$, we propose a novel approach which consists in using images from classes that will be encountered by the model during the future incremental steps. 
In detail, the objective of the initial incremental step is extended from solely learning the initial classes $\mathcal{Y}_1$ into learning $\mathcal{Y}_1 \cup \mathcal{Y}_\mathcal{S}$ where $\mathcal{Y}_\mathcal{S}=\bigcup_{i=2}^{T}\mathcal{Y}_i$ are the future classes. The feature extractor is jointly trained with a classifier on $\mathcal{D}_1 \cup \mathcal{S}$. At the end of the initial step, the classifier is restricted to the base classes $\mathcal{Y}_1$ by dropping the weights corresponding to the future classes from the auxiliary dataset $\mathcal{S}$, and the feature extractor is frozen. Then, the training procedure continues according to the baseline method for the remaining incremental steps. Our proposed method is detailed in Algorithm~\ref{alg:fpcil}.

\begin{algorithm}
\small
\DontPrintSemicolon
\caption{Future-Proof Class-Incremental Learning (FPCIL)}\label{alg:fpcil}
\textbf{Input:} Data-flow $\{\mathcal{D}_i\}_{i=1}^{T}$ and pre-trained generative model G. \;
\textbf{Ouput:} Models $\{\theta_i\}_{i=1}^{T}$ composed of $\{\Phi_i,\mathcal{W}_i \}$. \; \;

\For{$t \in \{1,2,...,T\}$}{
\If{$t = 1$}{
\tcp{Predict future classes based on initial classes using oracle or GPT}
$\mathcal{Y}_\mathcal{S} \leftarrow$ PredictFuture($\mathcal{Y}_1$) \; \;
\tcp{Generate synthetic samples of future classes}
$\mathcal{S} \leftarrow$ GenerateSynthetic(G, $\mathcal{Y}_\mathcal{S}$) \; \;

\tcp{Train the model using real data of the first step and synthetic data of future steps}
Train $\theta_1$ using $\mathcal{D}_1 \cup \mathcal{S}$ \; 
Freeze $\Phi_1$ and remove $\mathcal{Y}_\mathcal{S}$ weights from $\mathcal{W}_1$ \;
}
\Else
{
\tcp{Incrementally update the classifier using dedicated method (e.g. FeTrIL, FeCAM)}
Train $\mathcal{W}_t$ using $\mathcal{D}_t$ \;
}
}
\end{algorithm}

In the initial step, our proposed method FPCIL trains on both the initial training dataset $\mathcal{D}_1$ and an additional dataset $\mathcal{S}$ which contains training data for predicted future classes. As only the data for the current step are available in the Class-Incremental Learning setting, we propose to use synthetic data instead of the real ones. We use a pre-trained text-to-image diffusion model to generate synthetic images from the future classes during the initial step. This means that only the labels of the future classes are necessary. Compared to recent approaches for Incremental Learning relying on a pre-trained diffusion model \citep{Jodelet_2023_ICCV,chen2023diffusepast,kim2024sddgr} for generating synthetic samples of past classes, our proposed method aims at generating synthetic samples of future classes instead. By primarily using the synthetic data to train the feature extractor, our proposed method circumvents the impact of the distribution gap between real and synthetic samples generated by the diffusion model, making it applicable to Exemplar-Free Class-Incremental Learning.

While the requirement of having access to the label of future classes may seem counter to the concept of Class-Incremental Learning, we argue that it is practical and compatible with most real-world applications of Class-Incremental Learning. In general, during the developing a deep learning model meant for continuously accumulating new knowledge while achieving a given task in an ever-evolving environment, both the task and the environment are known. Therefore, it is reasonable to assume that it is possible, based on this initial knowledge of the environment and task, to predict their possible future evolution. However, acquiring beforehand real data corresponding to these future states of the environment and task may be difficult, expensive, or even impossible in some cases which lead us to use synthetic samples instead. As a concrete example, we propose to consider the case of an agent that would be deployed on an embedded device for assisting its user inside his home. The agent would have to be able to learn new concepts based on data directly collected from its environment depending on the particular need of its user. It would have to rely on incremental learning as frequent retraining from scratch would not be possible on either a cloud server as uploading user's data would be difficult due to privacy reasons, or in local due to the limited computational and memory capacity of the embedded device. Therefore, using our proposed method it is possible to train an initial model precisely tailored for each user (or group of users) before deployment, by relying on available knowledge about the user and its environment (interests, location, particular requests, etc.) to preemptively learn specific concepts that may appear in the future in addition to generic concepts. Since building a specific dataset of real samples for all possible needs of all possible users would not be tractable due to the significant cost in time and resources of the data collection process, the use of synthetic data appears as a faster and cheaper alternative. As a second example, we consider the case of an agent deployed on an embedded device for performing a task in a hard to access and potentially hostile environment where the agent has to be autonomous due to unreliable remote control possibility (deep sea, space, etc). The nature of the environment intrinsically limits the capacity to collect large training datasets in advance but its properties may be known. Using our proposed method, the agent could be trained before its deployment using synthetic data from generative diffusion models, or more realistically in this scenario using simulation-based tools. The agent would then be able to incrementally learn and adjust itself to the environment using real data collected while performing the task.

It should also be noted that based on our experiments detailed in the ablation study, the classes in the auxiliary dataset $\mathcal{S}$ do not need to be perfectly predicted to have a positive impact on the performance. For clarity, we refer to FPCIL as \enquote{FPCIL-Oracle} when the list of future classes has been perfectly predicted and \enquote{FPCIL-Partial (k\%)} otherwise, with $k$ the percentage of future classes correctly predicted.

Predictions of the future classes can be done through expert knowledge or through automated methods. Here, we introduce a simple approach for automatically predicting the future classes based on recently popular large language models. Our proposed method relies on the GPT model~\citep{brown2020language} \enquote{GPT-3.5 Turbo Instruct} used through the OpenAI API. The large language model is queried for a completion task using the prompt \enquote{The dataset contains the following 100 classes: $c_1$, [...], $c_k$,} where $c_1$ to $c_k$ are the names of the $k$ classes from the initial step, with $k$ equals to $\lvert \mathcal{Y}_1\rvert$. The number \enquote{100} is used here to encourage the model to generate a longer list of possible future classes as initial experiments without specifying a number would result in only a few number of guesses. In practice, the output of GPT has to be parsed to extract the predicted future classes. The completion function of GPT often continues the prompt as if it was part of a website or a scientific paper by appending imaginary details such as the number of images per class in the dataset, nonexistent download url for the dataset, or architecture of the model. Due to the random nature of the output from GPT, the prompt is queried 10 times and the predicted classes are ordered according to the number of times they have been predicted. The auxiliary dataset $\mathcal{S}$ is then restricted to the classes which were the most often predicted. We evaluated three versions with an increasing level of restriction: \enquote{FPCIL-GPT}, \enquote{FPCIL-GPT R1}, and \enquote{FPCIL-GPT R2}. Those approaches based on GPT also serve to imitate a more realistic application of Future-Proof Class-Incremental Learning by predicting various number of future classes with various accuracy.

\subsection{Generating future data}

\begin{figure*}
\begin{center}
\resizebox{\textwidth}{!}{%
\includegraphics{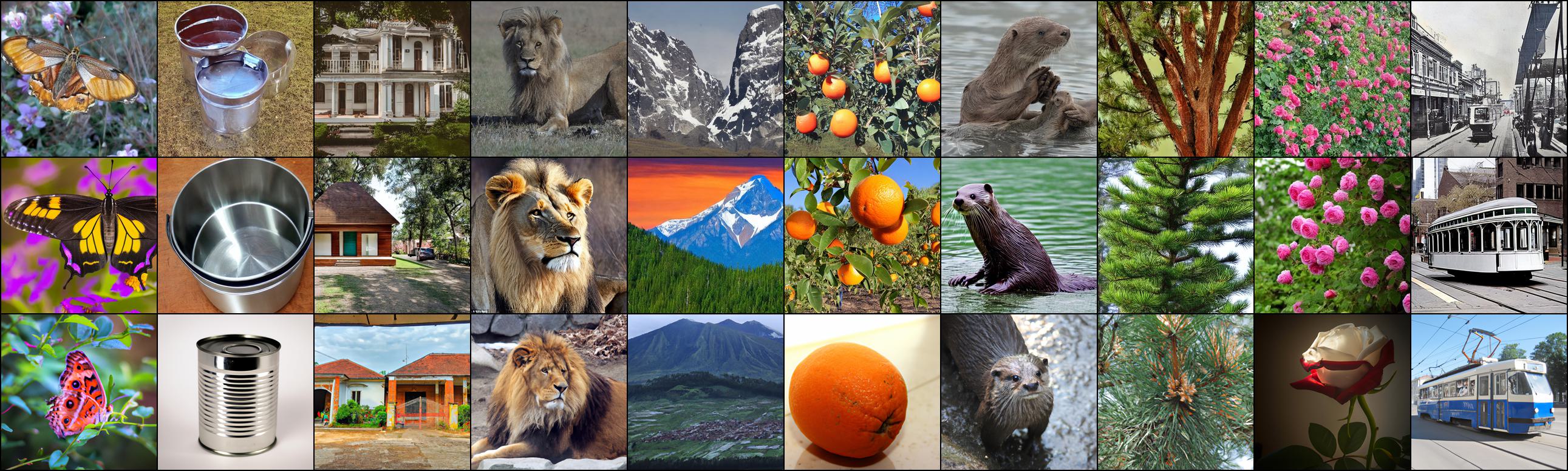}}
\end{center}
   \caption{Synthetic images of future classes for CIFAR100 generated using different diffusion models. Images are displayed before being resized for better appreciation. First row has been generated by Stable Diffusion 1.4 with a guidance scale of $2.0$, second row using  Stable Diffusion 1.4 with a guidance scale of $7.5$, and last row using DALL-E2. From left to right: butterfly, can, house, lion, mountain, orange, otter, pine tree, rose, and streetcar.}
\label{fig:diffgrid}
\end{figure*}

Leveraging recent advances in diffusion models and the wide availability of generative text-to-image models pre-trained on natural images, we propose to use them for generating the synthetic images dataset of future classes following the procedure initially defined by \citet{Sariyildiz_2023_CVPR}.

The auxiliary dataset $\mathcal{S}$ is generated using a pre-trained diffusion model $G$ conditioned solely on a textual prompt. As proposed in~\citep{Sariyildiz_2023_CVPR}, the default prompt used is \enquote{$c, d_c$} with \enquote{$c$} the name of the class and \enquote{$d_c$} a simple description of the class. Adding the description of the class into the prompt is intended to ensure that the generated images belong to the targeted class and not to any homonyms: for example \enquote{crane} in the ImageNet dataset, which may refer to either the bird or the machine used to move heavy objects. The definition \enquote{$d_c$} also tends to restrain the domain of the generated images even though the domain could also be enforced by adding it to the prompt. To automatically create the prompts, we rely on WordNet~\citep{10.1145/219717.219748} and use the lemmas of the synset as \enquote{$c$} and the definition of the synset as \enquote{$d_c$}. For each class, $n$ synthetic images are generated with $n$ equals to the number of images per class in the target dataset. Compared to traditional pre-training and concurrent method~\citep{bellitto2022effects}, our proposed method does not require the costly and fastidious process of collecting, curating, and annotating a dataset of real images. Examples of synthetic images generated for CIFAR100 are displayed in Figure~\ref{fig:diffgrid}.

The guidance scale is an important hyper-parameter of diffusion models relying on classifier-free guidance~\citep{ho2021classifierfree}. To some extent, it can be used to control the trade-off between the diversity and the quality of the generated synthetic images. For Stable Diffusion, the default recommended value for the guidance scale is often $7.0$ but following~\citep{Jodelet_2023_ICCV,Sariyildiz_2023_CVPR} we use the lower value of $2.0$ as a higher diversity is beneficial for pre-training.

While this generation procedure allows our proposed method to achieves significant improvement compared to baselines, there remains an important distribution gap between the real and synthetic data~\citep{Jodelet_2023_ICCV}. Reducing this distribution gap and its impacts on the training of the feature extractor would further improve the performance of our proposed method. As possible future research direction, the diffusion model could be finetuned using the data from the first incremental step, a more sophisticated prompting mechanism could be implemented, or a selection process could be used to discard the synthetic samples too far away from the distribution of the target dataset.

\begin{table*}
\caption{Definition of the different auxiliary datasets used during the first incremental step for each setting. \enquote{Class overlap} is the percentage of classes (actual number of classes in parentheses) from the future incremental steps of the target dataset that are in the auxiliary dataset $\mathcal{S}$, i.e. the number of correctly predicted future classes.}
\label{tab:datasetsdef}
\resizebox{\linewidth}{!}{%
    \begin{tabular}{l l l c c c c c}
    \hline
    Target dataset & Auxiliary dataset $\mathcal{S}$ & \makecell{Nature \\ of data} & Setting & \makecell{Number of \\ classes \\ $\lvert \mathcal{Y}_\mathcal{S} \rvert$} & \makecell{Samples \\ per class \\ $n$} & \makecell{Class overlap \\ with future steps \\ of target dataset} & Note \\
    \hline
    \multirow{10}{*}{\makecell{CIFAR100 \\ (100 classes)}}& CIFAR10 & Real & All & 10 & 5000 & 0\% & \\
    \cline{2-8}
    & \multirow{3}{*}{FPCIL-Partial (k\%) } & \multirow{3}{*}{Synthetic} & B50 Inc & 50 & 500 & k\% & \multirow{3}{0.75\linewidth}{k\% of the classes have been randomly selected from the target dataset, the remaining have been randomly selected from the wrong predictions of GPT-3.5. For k=100, we use the denomination \enquote{FPCIL-Oracle}.} \\
    & &  & B0 Inc10 & 90 & 500 & k\% & \\
    & &  & B0 Inc5 & 95 & 500 & k\% & \\
    \cline{2-8}
    & \multirow{2}{*}{FPCIL-GPT} & \multirow{2}{*}{Synthetic} & B0 Inc10 & 150 & 500 & 58.9\% (53) & \multirow{2}{0.75\linewidth}{Classes the most often predicted by GPT-3.5 based on the classes encountered in the first incremental step.} \\
    &  & & B0 Inc5 & 166 & 500 & 53.7\% (51) & \\
     \cline{2-8}
    & \multirow{2}{*}{FPCIL-GPT R1 } & \multirow{2}{*}{Synthetic} & B0 Inc10 & 90 & 500 & 47.8\% (43) & \multirow{2}{0.75\linewidth}{Subset of \enquote{FPCIL-GPT} containing a more restricted set of classes.} \\
    & & & B0 Inc5 & 102 & 500 & 33.7\% (32) & \\
     \cline{2-8}
    & \multirow{2}{*}{FPCIL-GPT R2 } & \multirow{2}{*}{Synthetic} & B0 Inc10 & 53 & 500 & 33.3\% (30) & \multirow{2}{0.75\linewidth}{Subset of \enquote{FPCIL-GPT} containing a more restricted set of classes. More restrictive than R1.} \\
    & & & B0 Inc5 & 58 & 500 & 21.0\% (20) & \\
     \hline
     \hline

     \multirow{16}{*}{\makecell{ImageNet-Subset \\ (100 classes)}} & \multirow{3}{*}{ImageNet-Compl.} & \multirow{3}{*}{Real} & B50 Inc & 50 & $\sim$1300 & 0\% & \multirow{3}{0.75\linewidth}{Classes have been randomly sampled from the remaining 900 classes of ILSVRC 2012. No class overlap with ImageNet-Subset.} \\
      &  &  & B0 Inc10 & 90 & $\sim$1300 & 0\% & \\
      &  &  & B0 Inc5 & 95 & $\sim$1300 & 0\% & \\
        \cline{2-8}
      & \multirow{3}{*}{ImageNet-Compl. X2} & \multirow{3}{*}{Real} & B50 Inc & 100 & $\sim$1300 & 0\% & \multirow{3}{0.75\linewidth}{Classes have been randomly sampled from the remaining 900 classes of ILSVRC 2012. No class overlap with ImageNet-Subset.}\\
      &  &  & B0 Inc10  & 180 & $\sim$1300 & 0\% & \\
      &  &  & B0 Inc5 & 190 & $\sim$1300 & 0\% & \\
      \cline{2-8}
      & \multirow{2}{*}{ImageNet-Compl. Animals} & \multirow{2}{*}{Real} & B0 Inc10  & 90 & $\sim$1300 & 0\% & \multirow{2}{0.75\linewidth}{Classes have been randomly sampled among the \enquote{animals} classes from the remaining 900 classes of ILSVRC 2012. No class overlap with ImageNet-Subset.} \\
      &  &  & B0 Inc5 & 95 & $\sim$1300 & 0\% & \\
      \cline{2-8}
      & \multirow{3}{*}{FPCIL-Oracle } & \multirow{3}{*}{Synthetic} & B50 Inc & 50 & 1300 &  100\% (50) & \multirow{3}{0.75\linewidth}{Future classes perfectly predicted.} \\
      &  &  & B0 Inc10  & 90 & 1300 & 100\% (90) & \\
      &  &  & B0 Inc5 & 95 & 1300 & 100\% (95) & \\
      \cline{2-8}
      & \multirow{3}{*}{FPCIL-Partial (0\%)} & \multirow{3}{*}{Synthetic} & B50 Inc & 50 & 1300 &  0\% & \multirow{3}{0.75\linewidth}{Same classes as \enquote{ImageNet-Compl.}.}\\
      &  &  & B0 Inc10  & 90 & 1300 &  0\% & \\
      &  &  & B0 Inc5 & 95 & 1300 &  0\% & \\
      \cline{2-8}
      & \multirow{2}{*}{FPCIL-Partial Animals (0\%)} & \multirow{2}{*}{Synthetic} & B0 Inc10  & 90 & 1300 &  0\% & \multirow{2}{0.75\linewidth}{Same classes as \enquote{ImageNet-Compl. Animals}.} \\
      &  &  & B0 Inc5 & 95 & 1300 &  0\% & \\
      \hline
     
    \end{tabular}}
\end{table*}

\section{Experiments}

\subsection{Experimental setup}
\subsubsection{Datasets}
The experiments are conducted on two large scale datasets: CIFAR100~\citep{krizhevsky2009learning} and ImageNet-Subset. CIFAR100 is a dataset containing 60,000 32$\times$32 RGB images evenly divided among 100 classes with 500 training images and 100 testing images per class. ImageNet-Subset is subset of the high-resolution images dataset ImageNet (ILSVRC 2012)~\citep{russakovsky2015ImageNet}. This subset contains 100 randomly selected classes with about 1,300 training images and 50 testing images per class. Following \citet{Rebuffi_2017_CVPR}, the classes are ordered using numpy with the random seed 1993.

Two training procedures are used: Training from Scratch (TFS) and Training From Half (TFH). For the Training from Scratch procedure, the classes are evenly divided into $T$ (10, or 20) incremental steps, each incremental step containing the same number of classes. For the Training From Half procedure, following~\citep{Hou_2019_CVPR}, the first step contains half of the classes and the remaining classes are evenly divided into the following $T-1$ incremental steps. Following \citet{zhou2023deep}, the two training procedures are referred to using the common naming system \enquote{B-$i$ Inc-$j$} with $i$ the number of classes in the first step and $j$ the number of classes each of the following incremental step; $i$ equals to zero is used to refer to the Learning From Scratch procedure.

\subsubsection{Model training}
Following common practice for Exemplar-Free Class-Incremental Learning~\citep{Petit_2023_WACV,Zhu_2021_CVPR,zhu2022self}, ResNet-18 is used for both CIFAR100 and ImageNet-Subset. The feature extractor is trained solely during the first incremental step and is frozen for the rest of the training procedure. The backbone is trained with SDG using the standard Softmax Cross-Entropy loss and AutoAugment~\citep{Cubuk_2019_CVPR} for 160 epochs with a batch size of 128 and a weight decay of $5e^{-4}$. The initial learning rate is set to 0.1 and decrease to zero using a cosine annealing schedule. For the classifier, we use a fully-connected linear layer trained following FeTrIL~\citep{Petit_2023_WACV}. The classifier is trained using SGD for 50 epochs at each incremental step, with an initial learning rate of 0.1 and cosine annealing schedule. In the ablation study, we also report results using a Nearest Class Mean (NCM) Classifier and FeCAM \citep{goswami2023fecam}. If not stated otherwise, the reported results for the different baselines have been reproduced by our own experiments.
Experiments are conducted using Pytorch~\citep{Paszke_PyTorch_An_Imperative_2019} and PyCIL~\citep{zhou2023pycil}.

\begin{table*}
\begin{center}
\caption{Performances on CIFAR100 with the Learning from Half and Learning From Scratch procedures. Synthetic samples generated using Stable Diffusion with a guidance scale of $2.0$ . Using ResNet-18 as the backbone. \enquote{AVG} is Average Incremental Accuracy (Top-1) and \enquote{Last} is the final overall accuracy of the model after the last incremental step. Best result is marked in bold and second best is underlined. Results averaged over 3 random runs.}
\label{tab:maincifar}

\resizebox{\textwidth}{!}{%
\begin{tabular}{l cc c cc c cc c cc c cc}
\hline
\multirow{3}{*}{Methods} & \multicolumn{8}{c}{CIFAR100 - LFH} & & \multicolumn{5}{c}{CIFAR100 - LFS} \\
\cline{2-9} \cline{11-15}
& \multicolumn{2}{c}{ B50 Inc10} & & \multicolumn{2}{c}{ B50 Inc5} & & \multicolumn{2}{c}{ B50 Inc2} & & \multicolumn{2}{c}{ B0 Inc10} & & \multicolumn{2}{c}{ B0 Inc5} \\
\cline{2-3} \cline{5-6} \cline{8-9} \cline{11-12} \cline{14-15}
& AVG & Last & & AVG & Last & & AVG & Last & & AVG & Last & & AVG & Last \\
\hline
FeTrIL~\citep{Petit_2023_WACV} & 68.05& 58.69 && 66.81 & 58.22 && 65.46 & 56.94 && 50.50 & 33.86 && 40.80 & 24.65 \\
\hline
FeTrIL w/ Aux. CIFAR10 & \underline{70.60} & 60.82 && \underline{69.34} & 60.03  && \underline{68.23} & 59.26 && 58.56 & 41.72 && 52.42 & 35.03 \\
\hline
FeTrIL w/ FPCIL-Oracle & \textbf{73.44} & \textbf{65.71} && \textbf{72.81} & \textbf{65.20}  && \textbf{72.40} & \textbf{64.76} && \textbf{68.09} & \textbf{55.43} && \textbf{62.92} & \textbf{49.67} \\
FeTrIL w/ FPCIL-Partial (0\%) & 70.25 & \underline{61.41} && 69.27 & \underline{60.56}  && 68.13 & \underline{59.67} && 62.63 & 47.71 && 57.46 & 42.71 \\
\hline
FeTrIL w/ FPCIL-GPT & - & - && - & - && - & - && \underline{67.18} & \underline{53.89} && \underline{62.15} & \underline{48.67} \\
FeTrIL w/ FPCIL-GPT R1 & - & - && - & - && - & - && 66.79 & 53.02 && 60.65 & 46.08 \\
FeTrIL w/ FPCIL-GPT R2 & - & - && - & - && - & - && 65.25 & 51.01 && 59.89 & 44.33 \\
\hline
\end{tabular}}
\end{center}
\end{table*}

\begin{table*}
\begin{center}
\caption{Performances on ImageNet-Subset with the Learning from Half and Learning From Scratch procedures. Synthetic samples generated using Stable Diffusion with a guidance scale of $2.0$ . Using ResNet-18 as the backbone. \enquote{AVG} is Average Incremental Accuracy (Top-1) and \enquote{Last} is the final overall accuracy of the model after the last incremental step. Best result is marked in bold and second best is underlined. Results averaged over 3 random runs.}
\label{tab:mainImagenetsub}

\resizebox{\textwidth}{!}{%
\begin{tabular}{l cc c cc c cc c cc c cc}
\hline
\multirow{3}{*}{Methods} & \multicolumn{8}{c}{ImageNet-Subset - LFH} & & \multicolumn{5}{c}{ImageNet-Subset - LFS} \\
\cline{2-9} \cline{11-15}
& \multicolumn{2}{c}{ B50 Inc10} & & \multicolumn{2}{c}{ B50 Inc5} & & \multicolumn{2}{c}{ B50 Inc2} & & \multicolumn{2}{c}{ B0 Inc10} & & \multicolumn{2}{c}{ B0 Inc5} \\
\cline{2-3} \cline{5-6} \cline{8-9} \cline{11-12} \cline{14-15}
& AVG & Last & & AVG & Last & & AVG & Last & & AVG & Last & & AVG & Last \\
\hline
FeTrIL~\citep{Petit_2023_WACV} & 72.97 & 64.49 && 72.01 & 63.54 && 71.02 & 62.19 && 51.64 & 34.53 && 39.64 & 22.42 \\
\hline
FeTrIL w/ Aux. ImageNet-Compl. & 75.19 & 67.81 && 74.08 & 66.96 && 73.25 & 65.71 && 70.62 & 59.53 && 68.14 & 57.45 \\
FeTrIL w/ Aux. ImageNet-Compl. X2 & \underline{76.33} & \underline{69.52} && \underline{75.30} & \underline{68.53} && \underline{74.69} & \underline{67.89} && \underline{73.96} & \underline{64.68} && \textbf{73.01} & \underline{62.98} \\
\hline
FeTrIL w/ FPCIL-Oracle & \textbf{77.21} & \textbf{71.80} && \textbf{76.81} & \textbf{71.51} &&\textbf{ 76.84} & \textbf{71.17} && \textbf{74.04} & \textbf{65.65} && \underline{72.64} & \textbf{64.29} \\
FeTrIL w/ FPCIL-Partial (0\%) & 74.90 & 67.27 && 73.93 & 66.37 && 73.18 & 65.25 && 68.89 & 57.21 && 65.94 & 55.09 \\
\hline
\end{tabular}}
\end{center}
\end{table*}

\subsubsection{Synthetic generation}
Two diffusion models are used for generating the synthetic samples during training: Stable Diffusion~\citep{Rombach_2022_CVPR} and DALL-E2~\citep{ramesh2022hierarchical}. If not specified otherwise, the default diffusion model is Stable Diffusion v1.4~\footnote{https://huggingface.co/CompVis/stable-diffusion-v1-4} used with a guidance scale of $2.0$ for 50 steps. Following~\citep{Sariyildiz_2023_CVPR}, \enquote{$c, \; d_c$} is used for the prompt. The same datasets of synthetic images are used for all the experiments. Those datasets have been generated prior to incremental learning experiments, at size 512$\times$512 using a fixed seed and contains 500 images per class for CIFAR100 and 1300 images per class for ImageNet-Subset. For CIFAR100, the synthetic images are resized to 32$\times$32 before being fed to the model. For DALL-E2, because the model is used through the official API of OpenAI, it is not possible to control the different hyper-parameters of the model; the default, unknown, values are used for the guidance scale, the random seed and the number of denoising step. A slightly different prompt is used for DALL-E2, \enquote{a photo of a $c, \; d_c$}, as the default one would generate unsatisfactory images for some classes. In order to comply with the moderation policy of the API, few definitions $d_c$ have been changed.

\subsubsection{Baselines}
As a comparison baseline, a dataset containing real images from classes similar yet different to the ones in the target dataset is used as auxiliary dataset $\mathcal{S}$ during the first incremental step. The purpose of this baseline is to compare our proposed method with the more standard pre-training approach. It is similar to the method proposed by \citet{bellitto2022effects} adapted to Exemplar-Free Class-Incremental Learning with a frozen feature extractor.
For CIFAR100, we use CIFAR10 for the auxiliary dataset as both datasets were created using the same annotation and curation methodology from the TinyImages Dataset~\citep{4531741}. However, while CIFAR10 contains more samples than the datasets generated using our proposed method FPCIL, it is important to note that CIFAR10 contains only 10 classes. For ImageNet-Subset, the auxiliary dataset \enquote{ImageNet-Compl.} was created by randomly selecting classes from the complement of ImageNet-Subset in ImageNet. \enquote{ImageNet-Compl.} contains the same number of classes as the datasets generated by our proposed method FPCIL and approximately the same number of samples. This is a strong baseline as both subsets are sampled from the same dataset and, while they do not contain the exact same classes, classes in both datasets are from the same categories and some are highly similar: for example both datasets contain different breeds of dogs or species of insects. We additionally use \enquote{ImageNet-Compl. X2} which contains two times more classes than ImageNet-Compl. We also report the performance of \enquote{FPCIL-Partial (k\%)} which generates a dataset of the same size as FPCIL-Oracle but with only $k$\% of the classes correctly predicted. For CIFAR100, this auxiliary dataset is built by randomly sampling $k$\% of the classes from FPCIL-Oracle and the remaining ones from the wrong prediction of GPT-3.5. For ImageNet-Subset, \enquote{FPCIL-Partial (0\%)} contains synthetic images belonging to the same classes as the ones in ImageNet-Compl.
The definition of all the different auxiliary datasets used are summarized in Table~\ref{tab:datasetsdef}.

\subsubsection{Evaluation} 
Models are evaluated and compared using both the \enquote{Final Accuracy} and the \enquote{Average Incremental Accuracy}~\citep{Rebuffi_2017_CVPR}. The Average Incremental Accuracy is defined as the average of the Top-1 accuracy of the model on the test dataset of all the classes learned so far, at the end of each incremental step including the first one.

\subsection{Results}
Average incremental accuracy and final accuracy of our proposed method FPCIL are reported for CIFAR100 and ImageNet-Subset in Table~\ref{tab:maincifar} and Table~\ref{tab:mainImagenetsub} for both the Learning From Scratch and Learning From Half procedures.

Our proposed approach FPCIL considerably improves the average incremental accuracy of the state-of-the-art method FeTrIL~\citep{Petit_2023_WACV}, from 5.39 percentage points (\emph{p.p.}) up to 22.12\emph{p.p.} on CIFAR10 and from 4.24\emph{p.p.} up to 33.00\emph{p.p.} on ImageNet-Subset. This improvement is notably higher for the more difficult Learning From Scratch procedure which is particularly challenging for FeTrIL due to the limited quantity of data available during the first incremental step. 
Compared to using an auxiliary dataset of real images from classes different from those in the target dataset, our method also achieves significantly higher accuracy. This highlights the fundamental advantage of our proposed method: while the standard pre-training procedure improves the generalization capability of the feature extractor by leveraging additional data belonging to classes different from those in the target dataset, our proposed method aims to learn a feature extractor specifically designed for the incoming task by relying on data from future classes instead. Even if those data from future classes are synthetic, it allows to learn a feature extractor more suitable for the future tasks resulting in a significantly higher accuracy compared standard pre-training on real data. Moreover, on ImageNet-Subset, our proposed method FPCIL even outperforms a two times larger auxiliary dataset of real images, demonstrating that our proposed algorithm is more data-efficient that standard pre-training. Furthermore synthetic datasets can be rapidly generated whereas auxiliary datasets of real data require a long and costly collection, curation, and annotation process.
Finally, even in the case where the future classes have not been accurately predicted, the auxiliary dataset of synthetic images performs on par with the standard pre-training approach relying on real datasets: for the Learning From Half procedure, the difference of average incremental accuracy between using ImageNet-Compl or its synthetic counterpart is at most 0.29\emph{p.p.}. This makes our proposed method beneficial in every setting.

On CIFAR100, our more practical approach FPCIL-GPT also reaches a higher accuracy than the baseline method FeTrIL yet slightly lower than the FPCIL-Oracle. On CIFAR100 in the Learning From Scratch procedure with 10 steps of 10 classes (B0 Inc10), the less restricted FPCIL-GPT predicted a total of 150 classes with 53 that actually are among the 90 future classes while the most restricted FPCIL-GPT R2 predicted only 53 classes with 30 among the 90 future classes. All the details regarding future class predictions using GPT are reported in Table~\ref{tab:datasetsdef}. While the more restricted version FPCIL-GPT R1 and R2 tend to predict a lower number of classes that are not among the future classes, the less restricted version predicts a higher number of future classes, achieving higher performances. There is a trade-off between the precision and recall in the prediction of future classes, while predicting more classes improve the performance it also increase the computational cost of the first incremental step due to the cost of generating synthetic images and training the feature extractor. In the ablation study, we further analyze the impact of the correctness of the prediction of future classes.

\begin{table*}
\footnotesize
\caption{Performances on CIFAR100 with the Learning From Scratch procedure depending on the ratio of correctly predicted future classes in the complementary dataset used during the initial step. Synthetics samples generated using Stable Diffusion with a guidance scale of $2.0$ . Using ResNet-18 as the backbone. \enquote{AVG} is Average Incremental Accuracy (Top-1) and \enquote{Last} is the final overall accuracy of the model after the last incremental step. Best result is marked in bold. Results averaged over 3 random runs.}
\label{tab:wrongcifar}
\begin{center}
\begin{tabular}{l cc}
\hline
\multirow{2}{*}{Methods} & \multicolumn{2}{c}{CIFAR100 - B0 Inc10} \\
\cline{2-3}
& AVG & Last \\
\hline
FeTrIL~\citep{Petit_2023_WACV} & 50.50 & 33.86 \\
\hline
FeTrIL w/ FPCIL-Oracle (100\%) & \textbf{68.09} & \textbf{55.43} \\
FeTrIL w/ FPCIL-Partial (66\%) & 66.06 & 52.21 \\
FeTrIL w/ FPCIL-Partial (50\%) & 64.79 & 50.76 \\
FeTrIL w/ FPCIL-Partial (33\%) & 64.32 & 49.90 \\
FeTrIL w/ FPCIL-Partial  (0\%) & 62.63 & 47.71 \\
\hline
\end{tabular}
\end{center}
\end{table*}

\begin{table*}
\footnotesize
\begin{center}
\caption{Performances on ImageNet-Subset with the Learning From Scratch procedure depending on the complementary dataset used during initial step. Synthetic samples generated using Stable Diffusion with a guidance scale of $2.0$ . Using ResNet-18 as the backbone. \enquote{AVG} is Average Incremental Accuracy (Top-1) and \enquote{Last} is the final overall accuracy of the model after the last incremental step. Best result is marked in bold. Results averaged over 3 random runs.}
\label{tab:imagenetdomain}
\begin{tabular}{l cc c cc}
\hline
\multirow{2}{*}{Methods} & \multicolumn{5}{c}{ImageNet-Subset - LFS} \\
& \multicolumn{2}{c}{B0 Inc10} & & \multicolumn{2}{c}{B0 Inc5} \\
\cline{2-3} \cline{5-6}
& AVG & Last && AVG & Last \\
\hline
FeTrIL~\citep{Petit_2023_WACV} & 51.64 & 34.53 && 39.64 & 22.42\\
FeTrIL w/ FPCIL-Oracle & \textbf{74.04} & \textbf{65.65} && \textbf{72.64} & \textbf{64.29} \\
\hline
FeTrIL w/ Aux. ImageNet-Compl. Animals & 68.74 & 57.87 && 64.09 & 54.29 \\
FeTrIL w/ Aux. ImageNet-Compl. & 70.62 & 59.53 && 68.14 & 57.45 \\
\hline
FeTrIL w/ FPCIL-Partial Animals (0\%)  & 66.08 & 54.97 && 61.63 & 51.11 \\
FeTrIL w/ FPCIL-Partial (0\%) & 68.89 & 57.21 && 65.94 & 55.09 \\
\hline
\end{tabular}
\end{center}
\end{table*}

\subsection{Ablation study}

\subsubsection{Accurately predicting the future}

In Table~\ref{tab:wrongcifar}, we compare the performance of our proposed methods combined with FeTrIL on CIFAR100 B0 Inc10 depending on the ratio of correctly predicted future classes in the auxiliary dataset used during the initial step. Experimentally, for a fixed dataset size, the average incremental accuracy increase with the correctness of the prediction of the future classes. This confirms the importance of predicting the future classes compared to the standard pre-training procedure which relies on real data of classes different from those in the target dataset that will be encountered by the continual learner in the future. Moreover, even in the case where no future class has been correctly predicted, our proposed method still improves the performance of the baseline approach as, in most cases, additional data help to learn a feature extractor with better generalization capability.

Additionally, in Table~\ref{tab:imagenetdomain} we report the accuracy for an alternative version of ImageNet-Compl and FPCIL-Partial named respectively \enquote{ImageNet-Compl. Animals} and \enquote{FPCIL-Partial Animals}. ImageNet-Compl Animals is built by sampling classes from the complement to ImageNet-Subset in ImageNet while restricting the classes to animals; FPCIL-Partial Animals contains the same classes that ImageNet-Compl Animals but the images are generated using a diffusion model. These two more specialized datasets achieve notably lower accuracy than their more general version. This further emphasizes the importance of predicting the future classes and not limiting the scope of the auxiliary dataset.

\subsubsection{Quality and diversity of synthetic samples}

\begin{table*}
\footnotesize
\begin{center}
\caption{Performances on CIFAR100 with the Learning from Half and Learning From Scratch procedures depending on the Diffusion Model used for generating the synthetic data: Stable Diffusion with a guidance scale of $2.0$, Stable Diffusion with a guidance scale of $7.5$, and DALL-E2. Using ResNet-18 as the backbone. \enquote{AVG} is Average Incremental Accuracy (Top-1) and \enquote{Last} is the final overall accuracy of the model after the last incremental step. Best result is marked in bold. Results averaged over 3 random runs.}
\label{tab:cifarDMs}
\begin{tabular}{l cc c cc c cc}
\hline
\multirow{3}{*}{Methods} & \multicolumn{8}{c}{CIFAR100} \\
\cline{2-9}
& \multicolumn{2}{c}{ B50 Inc5} & & \multicolumn{2}{c}{ B0 Inc10} & & \multicolumn{2}{c}{ B0 Inc5}\\
\cline{2-3} \cline{5-6} \cline{8-9}
& AVG & Last & & AVG & Last & & AVG & Last \\
\hline
FeTrIL~\citep{Petit_2023_WACV} & 66.81 & 58.22 && 50.50 & 33.86 && 40.80 & 24.65 \\
\hline
FeTrIL w/ FPCIL-Oracle  SD $g=2.0$ & \textbf{72.81} & \textbf{65.20} && \textbf{68.09} & \textbf{55.43} && \textbf{62.92} & \textbf{49.67} \\
FeTrIL w/ FPCIL-Oracle  SD $g=7.5$ & 72.22 & 64.40 && 66.48 & 53.36 && 61.63 & 48.50 \\
FeTrIL w/ FPCIL-Oracle  DALL-E2 & 71.08 & 63.02 && 63.94 & 50.39 && 59.76 & 46.34 \\
\hline
\end{tabular}
\end{center}
\end{table*}

\begin{table*}
\begin{center}
\caption{Performances on CIFAR100 with the Learning from Half and Learning From Scratch procedures for FeTrIL depending on the number of synthetic samples $n$ generated for each future class. Synthetic samples generated using Stable Diffusion with a guidance scale of $2.0$ . Using ResNet-18 as the backbone. \enquote{AVG} is Average Incremental Accuracy (Top-1) and \enquote{Last} is the final overall accuracy of the model after the last incremental step. Results averaged over 3 random runs.}
\label{tab:cifarnbsamples}
\footnotesize
\begin{tabular}{l cc c cc c cc}
\hline
\multirow{3}{*}{Methods} & \multicolumn{8}{c}{CIFAR100} \\
\cline{2-9}
& \multicolumn{2}{c}{ B50 Inc5} & & \multicolumn{2}{c}{ B0 Inc10} & & \multicolumn{2}{c}{ B0 Inc5}\\
\cline{2-3} \cline{5-6} \cline{8-9}
& AVG & Last & & AVG & Last & & AVG & Last \\
\hline
FeTrIL~\citep{Petit_2023_WACV} & 66.81 & 58.22 && 50.50 & 33.86 && 40.80 & 24.65 \\
FeTrIL w/ FPCIL-Oracle $n=50$ & 68.31 & 59.67 && 60.45 & 45.56 && 55.52 & 40.89 \\
FeTrIL w/ FPCIL-Oracle $n=250$ & 71.27 & 63.23 && 65.53 & 52.40 && 61.45 & 48.38 \\
FeTrIL w/ FPCIL-Oracle $n=500$ & 72.81 & 65.20 && 68.09 & 55.43 && 62.92 & 49.67 \\
FeTrIL w/ FPCIL-Oracle $n=1000$ & 73.95 & 66.77 && 70.24 & 58.37 && 64.89 & 52.73 \\

\hline
\end{tabular}
\end{center}
\end{table*}

In Table~\ref{tab:cifarDMs}, the performance of FPCIL Oracle is reported on CIFAR100 depending on the diffusion model used to generate the synthetic images of future classes. By default in our study, we used Stable Diffusion with a guidance scale of $2.0$~\citep{Sariyildiz_2023_CVPR} to favor the diversity to the detriment of the quality. We verified this assumption by comparing it against Stable Diffusion with a higher guidance scale and DALL-E2 which both tend to favor more the quality than the diversity. Experiments confirm that when pre-training the feature extractor on future classes, the diversity is more important than the quality of the generated samples: Stable Diffusion with a small guidance scale achieves the highest results, ahead of DALL-E2 and Stable Diffusion with the default guidance scale.

In Table~\ref{tab:cifarnbsamples}, the performance of FPCIL Oracle is reported on CIFAR100 for different number of synthetic samples $n$ generated for each future classes. While the default value of $n$ equals 500 was used in the other experiments in order to match the size of the target dataset, it appears that just 50 samples per class is enough to improve the performance of the baseline and even outperform the use of eternal dataset of unrelated classes such as CIFAR10 which is more than 10 times bigger. Moreover, it is possible to further improve the performance of our proposed method by generating more synthetic data if the constraints of the system permits it.

In addition to the trade-of between quality and diversity, it is important to emphasize that a significant gap remains between real images and the synthetic samples generated using pre-trained diffusion models. As illustrated in Table~\ref{tab:imagenetdomain}, while Aux. ImageNet-Compl. and FPCIL-Partial (0\%) contain the same classes and almost the same number of samples, there is a difference of about 2 percentage points of accuracy between the feature extractor trained on those datasets. The exact same observation can be made for Aux. ImageNet-Compl. Animals and FPCIL-Partial Animals (0\%). Similarly, while achieving significantly higher results than the baseline, FPCIL-Oracle still achieves lower performances on both CIFAR100 and ImageNet-Subset compared to a upper-bound model which would have been trained on the complete target dataset at once.

\begin{table*}
\begin{center}
\caption{Performances on CIFAR100 with the Learning from Half and Learning From Scratch procedures for FeTrIL depending on the nature (real or synthetic) of the data used to train the feature extractor during the first incremental step. Synthetic samples generated using Stable Diffusion with a guidance scale of $2.0$ . Using ResNet-18 as the backbone. \enquote{AVG} is Average Incremental Accuracy (Top-1) and \enquote{Last} is the final overall accuracy of the model after the last incremental step. Best result is marked in bold. Results averaged over 3 random runs.}
\label{tab:cifarablanature}
\footnotesize
\begin{tabular}{l cc c cc c cc c cc}
\hline
\multirow{3}{*}{Methods} & & & & \multicolumn{8}{c}{CIFAR100} \\
\cline{5-12}
& \multirow{2}{*}{\makecell{Initial \\ classes}} & \multirow{2}{*}{\makecell{Future \\ classes}} & & \multicolumn{2}{c}{ B50 Inc5} & & \multicolumn{2}{c}{ B0 Inc10} & & \multicolumn{2}{c}{ B0 Inc5}\\
\cline{5-6} \cline{8-9} \cline{11-12}
& & & & AVG & Last & & AVG & Last & & AVG & Last \\
\hline
FeTrIL~\citep{Petit_2023_WACV} & real & - & & 66.81 & 58.22 && 50.50 & 33.86 && 40.80 & 24.65 \\
FeTrIL w/ FPCIL-Oracle var1 & - & synth. & & 47.00 & 44.10 && 64.05 & 53.30 && 64.07 & 52.44 \\
FeTrIL w/ FPCIL-Oracle var2 & synth. & synth. & & 58.04 & 53.52 && 65.56 & 54.42 && \textbf{65.24} & \textbf{53.51} \\
FeTrIL w/ FPCIL-Oracle & real & synth. & & \textbf{72.81} & \textbf{65.20} && \textbf{68.09} & \textbf{55.43} && 62.92 & 49.67 \\
\hline
\end{tabular}
\end{center}
\end{table*}

\begin{table*}
\begin{center}

\footnotesize
\caption{Performances on CIFAR100 ~\protect\subref{tab:backbone:lfh}~B50 Inc5 and ~\protect\subref{tab:backbone:lfs}~B0 Inc10 depending on the backbone used. Synthetic samples generated using Stable Diffusion with a guidance scale of $2.0$ . \enquote{AVG} is Average Incremental Accuracy (Top-1) and \enquote{Last} is the final overall accuracy of the model after the last incremental step. Best result is marked in bold. Results averaged over 3 random runs.}
\label{tab:backbone}

\subfloat[CIFAR100 - B50 Inc5\label{tab:backbone:lfh}]{%
\begin{tabular}{l cc c cc}
\hline
\multirow{2}{*}{Methods} & \multicolumn{5}{c}{CIFAR100 - B50 Inc5} \\
& \multicolumn{2}{c}{ResNet-32} & & \multicolumn{2}{c}{ResNet-18} \\
\cline{2-3} \cline{5-6}
& AVG & Last && AVG & Last \\
\hline
FeTrIL~\citep{Petit_2023_WACV} & 64.27 & 55.19 && 66.81 & 58.22 \\
\hline
FeTrIL w/ Aux. CIFAR10 & 64.48 & 55.85 && 69.34 & 60.03\\
\hline
FeTrIL w/ FPCIL-Oracle & \textbf{66.24} & \textbf{58.88} && \textbf{72.81} & \textbf{65.20} \\
FeTrIL w/ FPCIL-Partial (0\%) & 63.80 & 55.67 && 69.27 & 60.56\\
\hline
\end{tabular}}

\vspace{0.5cm}

\subfloat[CIFAR100 - B0 Inc10\label{tab:backbone:lfs}]{%
\begin{tabular}{l cc c cc}
\hline
\multirow{2}{*}{Methods} & \multicolumn{5}{c}{CIFAR100 - B0 Inc10} \\
& \multicolumn{2}{c}{ResNet-32} & & \multicolumn{2}{c}{ResNet-18} \\
\cline{2-3} \cline{5-6}
& AVG & Last && AVG & Last \\
\hline
FeTrIL~\citep{Petit_2023_WACV} & 46.32 & 29.80 && 50.50 & 33.86 \\
\hline
FeTrIL w/ Aux. CIFAR10 & 56.70 & 39.42 && 58.56 & 41.72 \\
\hline
FeTrIL w/ FPCIL-Oracle & \textbf{62.73} & \textbf{48.96} && \textbf{68.09} & \textbf{55.43} \\
FeTrIL w/ FPCIL-Partial (0\%) & 59.68 & 45.30 && 62.63 & 47.71 \\
\hline
\end{tabular}}
\end{center}
\end{table*}

\subsubsection{Combining real and synthetic data}

During the first incremental step, our proposed method FPCIL combines the real samples of the initial classes with synthetic samples of possible future classes to train the feature extractor. In Table~\ref{tab:cifarablanature}, different variations of our proposed method are compared on CIFAR100 to estimate the contribution of each component. \enquote{FPCIL var1} does not use the real samples from the first classes and trains the feature extractor solely using synthetic samples of future classes. \enquote{FPCIL var2} also generates synthetic samples of the initial classes and train the feature extractor using synthetic samples of both initial and future classes. In both cases, the classifier is then retrained at the end of the first incremental step using the real samples of the initial classes. Our proposed approach achieves the best performance overall, confirming the importance of using the available real data and not relying only on synthetic data. Interestingly, when the number of real data in the first incremental step is extremely small, replacing them by synthetic samples improves the overall performance while decreasing the final accuracy on the first classes. This may be due to bias induced in the learned representation by the distribution gap between real and synthetic samples being exacerbated in the most extreme settings.

\subsubsection{Architecture of feature extractor}

Table~\ref{tab:backbone} compares the performance of our method for two different feature extractors: the 11.2 million parameters ResNet-18 and the 0.46 million parameters ResNet-32. Due to the addition of the auxiliary dataset to the often limited dataset of the initial step, our method takes advantage of a larger feature extractor and results in a large improvement with the baseline compared to backbones with less parameters. Nonetheless, FPCIL remains competitive with smaller deep neural networks. 

\subsubsection{Pre-trained model and finetuning}

\begin{table*}
\begin{center}
\caption{Performances on CIFAR100 with the Learning from Half and Learning From Scratch procedures for FeTrIL using a ResNet-18 feature extractor pre-trained on ImageNet. Synthetic samples generated using Stable Diffusion with a guidance scale of $2.0$ . Using ResNet-18 as the backbone. Input data are upscaled before being fed to the model in order to match the pre-training configuration. \enquote{AVG} is Average Incremental Accuracy (Top-1) and \enquote{Last} is the final overall accuracy of the model after the last incremental step. Best result is marked in bold. Results averaged over 3 random runs.}
\label{tab:cifarpretrained}
\footnotesize
\begin{tabular}{l cc c cc c cc}
\hline
\multirow{3}{*}{Methods} & \multicolumn{8}{c}{CIFAR100} \\
\cline{2-9}
& \multicolumn{2}{c}{ B50 Inc5} & & \multicolumn{2}{c}{ B0 Inc10} & & \multicolumn{2}{c}{ B0 Inc5}\\
\cline{2-3} \cline{5-6} \cline{8-9}
& AVG & Last & & AVG & Last & & AVG & Last \\
\hline
FeTrIL~\citep{Petit_2023_WACV} & 66.73 & 57.98 && 51.05 & 34.43 && 42.71 & 26.19 \\
FeTrIL w/ Pre-training & 64.16 & 59.85 && 71.05 & 60.97 && 70.70 & 59.81 \\
FeTrIL w/ Pre-training + Finetune & 76.68 & 70.55 && 66.50 & 54.73 && 63.05 & 51.77 \\
FeTrIL w/ FPCIL-Oracle & 73.59 & 65.83 && 68.10 & 55.20 && 63.38 & 50.54 \\
FeTrIL w/ Pre-training + FPCIL-Oracle & \textbf{79.06} & \textbf{73.87} && \textbf{76.60} & \textbf{67.14} && \textbf{74.83} & \textbf{64.30} \\
\hline
\end{tabular}
\end{center}
\end{table*}

In addition to the main experiments in which the feature extractor is trained from scratch, we also conducted experiments with a pre-trained feature extractor. Using a feature extractor pre-trained on a large external dataset is an effective approach when the low number of classes in the initial incremental step is not enough to train the feature extractor from scratch.

Compared to experiments where the feature extractor is trained from scratch, when using a pre-trained model the initial learning rate for the feature extractor is set to 0.001 and the number of epochs is reduced to 80, other hyper-parameters are kept unchanged. In Table~\ref{tab:cifarpretrained}, the performances of FeTrIL on CIFAR100 using both a ResNet-18 feature extractor trained from scratch during the first incremental step and a ResNet-18 pre-trained on ImageNet.
While it appears that slightly finetuning the feature extractor using the data from the initial incremental step is detrimental if the number of classes in too small, it may also further improve the performance when enough classes are available in the first incremental step, especially if there is a large domain gap between the target dataset and the pre-training dataset. In comparison, FPCIL can be combined with pre-training independently of the number of classes in the first incremental step, increasing the average incremental accuracy by 2.38\emph{p.p.} up to 5.55\emph{p.p.}.This highlights the versatility of our proposed method FPCIL which can be used either for training the feature extractor from scratch or for finetuning a feature extractor pre-trained on a large dataset, significantly improving the performance of the baseline in both cases.

\subsubsection{Comparison with additional baselines}

\begin{table*}
\begin{center}
\caption{Performances on CIFAR100 with the Learning from Half and Learning From Scratch procedures for FeTrIL, FeCAM, and Nearest Class Mean (NCM) Classifier. Synthetic samples generated using Stable Diffusion with a guidance scale of $2.0$ . Using ResNet-18 as the backbone. \enquote{AVG} is Average Incremental Accuracy (Top-1) and \enquote{Last} is the final overall accuracy of the model after the last incremental step. Best result is marked in bold. Results averaged over 3 random runs.}
\label{tab:cifarmethods}
\footnotesize
\begin{tabular}{l cc c cc c cc}
\hline
\multirow{3}{*}{Methods} & \multicolumn{8}{c}{CIFAR100} \\
\cline{2-9}
& \multicolumn{2}{c}{ B50 Inc5} & & \multicolumn{2}{c}{ B0 Inc10} & & \multicolumn{2}{c}{ B0 Inc5}\\
\cline{2-3} \cline{5-6} \cline{8-9}
& AVG & Last & & AVG & Last & & AVG & Last \\
\hline
NCM  & 65.24 & 55.23 && 40.48 & 23.37 && 30.41 & 15.00 \\
\rowcolor{lightgray!50}
NCM w/ FPCIL-Oracle  & 71.61 & 63.21 && 60.98 & 46.39 && 55.69 & 41.10 \\
\rowcolor{lightgray!50}
\quad Improvement in \emph{p.p.} & +6.37& +7.98 && +20.50& +23.02&& +25.28& +26.10\\
\hline
FeTrIL~\citep{Petit_2023_WACV} & 66.81 & 58.22 && 50.50 & 33.86 && 40.80 & 24.65 \\
\rowcolor{lightgray!50}
FeTrIL w/ FPCIL-Oracle & 72.81 & 65.20 && 68.09 & 55.43 && 62.92 & 49.67 \\
\rowcolor{lightgray!50}
\quad Improvement in \emph{p.p.} & +6.00 & +6.98 && +17.59 & +21.57 && +22.12 & +25.02   \\
\hline
FeCAM~\citep{goswami2023fecam} & 70.79 & 63.04 && 52.85 & 38.47 && 43.30 & 28.88 \\
\rowcolor{lightgray!50}
FeCAM w/ FPCIL-Oracle & \textbf{76.77} & \textbf{70.49} && \textbf{71.14} & \textbf{59.95} && \textbf{68.31} & \textbf{56.05}  \\
\rowcolor{lightgray!50}
\quad Improvement in \emph{p.p.} & +5.98& +7.45 && +18.29& +21.48&& +25.01& +27.17\\
\hline
\end{tabular}
\end{center}
\end{table*}

\begin{table*}
\begin{center}
\caption{Performances on CIFAR100 with the Learning From Scratch procedures. Synthetic samples generated using Stable Diffusion with a guidance scale of $2.0$ . Using ResNet-32 as the backbone for CIFAR100 and ResNet-18 for ImageNet-Subset. Results marked with $\dagger$ are reported from \citep{meng2024diffclass}. \enquote{AVG} is Average Incremental Accuracy (Top-1) and \enquote{Last} is the final overall accuracy of the model after the last incremental step. Best result is marked in bold. Results averaged over 3 random runs.}
\label{tab:efcilcomparison}
\footnotesize
\begin{tabular}{l cc c cc c cc c cc}
\hline
\multirow{3}{*}{Methods} & \multicolumn{5}{c}{CIFAR100} & & \multicolumn{5}{c}{ImageNet-Subset} \\
\cline{2-6} \cline{8-12}
&  \multicolumn{2}{c}{ B0 Inc10} & & \multicolumn{2}{c}{ B0 Inc5} & &  \multicolumn{2}{c}{ B0 Inc10} & & \multicolumn{2}{c}{ B0 Inc5}\\
\cline{2-3} \cline{5-6} \cline{8-9} \cline{11-12} 
& AVG & Last & & AVG & Last & & AVG & Last & & AVG & Last \\
\hline
DiffClass$\dagger$~\citep{meng2024diffclass} & \textbf{68.05} & \textbf{58.40} && \textbf{67.10} & \textbf{57.11} && 73.87 & 67.02 && 72.51 & 68.68  \\
\hline
FeTrIL~\citep{Petit_2023_WACV} & 46.32 & 29.80 && 37.75 & 21.26 &&  51.64 & 34.53 && 39.64 & 22.42 \\
\rowcolor{lightgray!50}
FeTrIL w/ FPCIL-Oracle & 62.73 & 48.96 && 59.89 & 46.62 && 74.04 & 65.65 &&  72.64 & 64.29 \\
\hline
FeCAM~\citep{goswami2023fecam} & 44.12 & 27.83 && 35.57 & 20.04 && 55.37 & 40.03 && 43.48 & 29.11  \\
\rowcolor{lightgray!50}
FeCAM w/ FPCIL-Oracle & 64.47 & 52.01 && 63.25 & 50.00 && \textbf{77.78} & \textbf{69.85} && \textbf{77.41} & \textbf{69.17}  \\

\hline
\end{tabular}
\end{center}
\end{table*}

Additionally to FeTrIL~\citep{Petit_2023_WACV}, we report in Table~\ref{tab:cifarmethods} the performance of our method combined with a Nearest Class Mean (NCM) Classifier and FeCAM~\citep{goswami2023fecam}. NCM is a simple yet effective method~\citep{janson2022a,ostapenko2022continual} for Class-Incremental Learning with a fixed feature extractor that does not require memory. FeCAM is a recently proposed approach for Exemplar-Free Class-Incremental Learning achieving new state-of-the-art performances. Similarly to FeTrIL, the performance of both NCM and FeCAM is highly dependent on the quality of the features produced by the backbone. It struggles with the Learning From Scratch procedure where the number of classes and data samples used to train the feature extractor is limited. Our method improves the performance of the NCM classifier and FeCAM in every considered setting and achieves the most significant improvements with the Learning From Scratch procedure on CIFAR100 B0 Inc5, reaching a final accuracy up to two times higher than the baseline approach.

Furthermore, in Table~\ref{tab:efcilcomparison} we compare our approach with the concurrently released method DiffClass~\citep{meng2024diffclass} which leverages the Stable Diffusion model for generating samples from past classes. While our method takes advantage of a larger backbone and can achieve higher performances as detailed in the ablation study, the comparison on CIFAR100 is done using ResNet-32 for the backbone as the results for DiffClass using ResNet-18 are not available. Experimentally, DiffClass achieves higher performance on CIFAR100 but is outperformed by our approach on ImageNet-Subset. It should be noted that \citet{meng2024diffclass} proposed several new components to mitigate the distribution gap between real and synthetic samples~\citep{Jodelet_2023_ICCV}. Some of those contributions, such as a multi-distribution matching technique to finetune the diffusion model and a selective synthetic image augmentation technique, could be adapted and combined with our proposed approach FPCIL to further improve its performance. 
Finally, our proposed method FPCIL conducts the computationally expensive task of generating synthetic samples using diffusion model only during the first incremental step, before freezing the feature extractor and only updating the classifier afterward. This allows us to potentially conduct the first incremental step on specialised hardware in a data center before deploying the model on devices with limited hardware where each following incremental step could be conducted in a matter of minutes on a CPU using approaches such as FeTrIL or FeCAM. In comparison, DiffClass finetunes the diffusion model, generates synthetic samples of past classes, and finetunes the whole feature extractor at each incremental step, resulting in much larger computational cost and requiring specialised hardware during whole training procedure. This makes our approach preferable if applicable.

\section{Conclusion}
In this work, we introduced a new method for Exemplar-Free Class-Incremental Learning which leverages large pre-trained diffusion model to generate images from future classes. Experimental results highlight how our method can significantly improve the accuracy of state-of-the-art approaches while only modifying the initial step. We found that our method requires less additional data than traditional methods relying on real curated datasets. While we focused our present study to feature extractors trained from scratch during the first incremental step, in future work, we will further investigate how synthetic images of future classes can be used to adapt a general pre-trained foundation model. Moreover, we will also explore how synthetic data of future classes can be leveraged during the whole training procedure, not only during the first incremental step.

\backmatter

\bmhead{Acknowledgements}

This work is partly supported by JSPS Grant-in-Aid for Scientific Research (grant number 23H03451, 21K12042) and the New Energy and Industrial Technology Development Organization (Grant Number JPNP20017).

\bmhead{Credit Author Statement}
Quentin Jodelet: Conceptualization, Investigation, Software, Writing - Original Draft. Xin Liu: Conceptualization, Supervision, Resources, Funding acquisition, Writing - Review \& Editing. Yin Jun Phua: Conceptualization, Supervision, Writing - Review \& Editing. Tsuyoshi Murata: Supervision, Resources, Funding acquisition, Writing - Review \& Editing.

\bmhead{Data Availability}

Benchmark datasets CIFAR-100, and ImageNet are publicly available. Diffusion models used for generating synthetic datasets are either publicly available or behind a paid API.

\section*{Declarations}

\bmhead{Conﬂict of interest}
The authors declare that they have no known competing financial interests or personal relationships that could have appeared to influence the work reported in this paper.

\bibliography{egbib}

\end{document}